\documentclass[10pt,twocolumn,letterpaper]{article}

\usepackage{mpi}
\usepackage{times}
\usepackage{epsfig}
\usepackage{graphicx}
\usepackage{amsmath}
\usepackage{amssymb}

\usepackage{comment}
\usepackage{mathrsfs} 

\usepackage{enumitem}

\usepackage{textcomp}

\usepackage{multirow}

\usepackage{color} 
\usepackage[table]{xcolor}
\usepackage{url}

\newcommand{\norm}[1]{\left\lVert#1\right\rVert}

\def\bfF{{\bf F}} 
\def\bfB{{\bf B}} 
\def\bfE{{\bf E}} 
\def\bfx{{\bf x}} 
\def\bfy{{\bf y}} 
\def\bfP{{\bf P}} 
\def\bfS{{\bf S}} 
\def\bfR{{\bf R}}
\def\bft{{\bf t}}
\def\bfI{{\bf I}} 
\def\bfM{{\bf M}} 
\def\bfQ{{\bf Q}} 
\def\bfC{{\bf C}} 
 
\def\bfX{{\bf X}} 
\def\bfY{{\bf Y}}

\def\bfq{{\bf q}}

\def\bfF{{\bf F}}
\def\bfE{{\bf E}}
\def\bfR{{\bf R}}

\def\bfS{{\bf S}}

\def\bfP{{\bf P}}

\def\bfB{{\bf B}}
\def\bfI{{\bf I}}
\def\bfC{{\bf C}}

\def\bfE{{\bf E}}

\def\bfx{{\bf x}}
\def\bfy{{\bf y}}

\def\bfF{{\bf F}}
\def\bfX{{\bf X}}
\def\bfY{{\bf Y}}

\def\bft{{\bf t}}

\def\Ham{{\boldsymbol{\hat{\mathcal{H}}}}}

\newlength{\bracewidth}

\usepackage[labelfont=bf,font={footnotesize}]{caption} 

\usepackage{array}

\usepackage[pagebackref=true,breaklinks=true,letterpaper=true,colorlinks,bookmarks=false]{hyperref} 

\mpifinalcopy

\begin{document}

\title{\vspace{-18pt}\hspace{-2pt}\hbox{A Quantum Computational Approach to Correspondence Problems on Point Sets}}

\author{ 
Vladislav Golyanik\hspace{3.5em} Christian Theobalt\\ 
Max Planck Institute for Informatics, Saarland Informatics Campus 
}

\maketitle 

\begin{abstract} 
  Modern adiabatic quantum computers (AQC) are already used to solve difficult combinatorial optimisation problems in various domains of science. 
  Currently, only a few applications of AQC in computer vision have been demonstrated. 
  We review AQC and derive a new algorithm for correspondence problems on point sets suitable for execution on AQC. 
  Our algorithm has a subquadratic computational complexity of the state preparation. 
  Examples of successful transformation estimation and point set alignment by simulated sampling are shown in the systematic experimental evaluation. 
  Finally, we analyse the differences in the solutions and the corresponding energy values. 
\end{abstract} 

\vspace{-7pt}

\section{Introduction}\label{sec:introduction}

Since their proposal in the early eighties \cite{Benioff1980, Manin1980, Feynman1982}, quantum computers have attracted much attention of physicists and computer scientists. 
Impressive advances both in quantum computing hardware and algorithms have been demonstrated over the last thirty years \cite{Lanting2014etal, Grover1996, Shor1997, Rebentrost2013, Lloyd2013, Coles2018etal, Devitt2016, Neukart2017, YLi2018, Nemoto2014}. 
Quantum computers are not universally faster than conventional machines, but they can natively execute algorithms relying on quantum parallelism, \textit{i.e.,} 
\textit{the ability to perform operations on exponentially many superimposed memory states simultaneously} \cite{Rieffel2000}. 
To harness the advantages, carefully designed algorithms are required. 
Nowadays, the motivation to take advantage of quantum effects in computing is also facilitated by the classical computing paradigm approaching its limits, 
since the quantum effects are becoming non-neglectable while manufacturing and using conventional CPUs. 
As a result, alternative paradigms such as massively parallel computing devices have been brought into being. 
While universal gate quantum computer technology has not yet reached the maturity, modern adiabatic quantum annealers (AQA) are already capable of solving difficult real-world combinatorial optimisation problems \cite{Bian2013, Bian2016, Denchev2016, Neukart2017}. 
The primary difference of universal gate quantum computing and AQA is that the latter can address objectives formulated as \textit{quadratic unconstrained binary  optimisation problems} (QUBOP) defined as 
\begin{equation}\label{eq:QUBOP} 
  \arg \min_{\bfq \in \bfB^n} \bfq^\mathsf{T} \bfP \bfq, 
\end{equation} 
where $\bfq$ is a set of $n$ binary variables, and $\bfP$ is a symmetric matrix of weights between the variables. 
The operational principle of AQA is grounded on the \textit{adiabatic theorem of quantum mechanics} \cite{Born1928} which states that 
\begin{equation}\label{th:adiabatic_theorem} 
  \hspace{-2pt}
  \begin{minipage}{0.9\columnwidth}
     \textit{if a quantum-mechanical system is in the ground state of a time-dependent Hamiltonian and parameters of this Hamiltonian are changing gradually enough, the system will continue to remain in the ground state during the evolution (\emph{see Table~\ref{tab:quantum_and_classic_notions} for quantum notions}).} 
  \end{minipage}
\end{equation}

\begin{figure}[t!] 
\centering 
\includegraphics[width=1.0\linewidth]{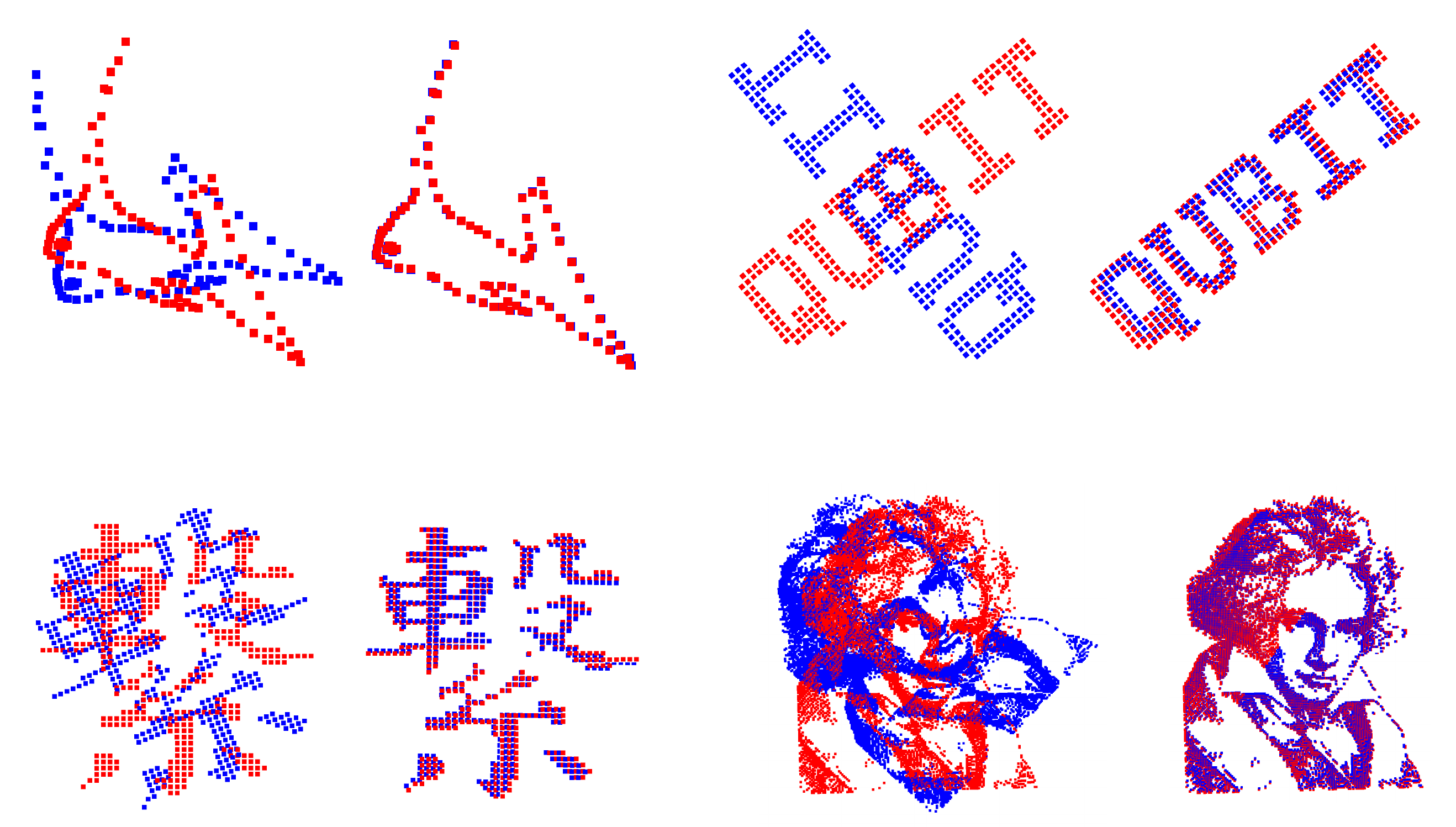} 
\caption{Different 2D point sets --- \textit{fish} \cite{Myronenko2010}, \textit{qubit}, \textit{kanji} and \textit{composer} --- aligned with our QA approach. 
For every pair of point sets, the initial misalignment is shown on the left, and the registration is shown on the right. 
QA is the first transformation estimation and point set alignment method which can be executed on adiabatic quantum computers. 
} %
\label{fig:DIFFERENT_2D_} 
\end{figure}

In their seminal paper, Farhi \textit{et al.}~\cite{Farhi2001} have shown that the adiabatic principle \eqref{th:adiabatic_theorem} can be used for 
solving $\mathcal{NP}$-complete optimisation problems and laid the foundation for adiabatic quantum computing. 
Several years later, Aharonov~\textit{et al.}~\cite{Aharonov2007} theoretically showed the equivalence between classical quantum computing 
and quantum annealing models. 
As of 2019-2020, general-purpose quantum computers accessible for research purposes and applications contain up to $20$ qubits \cite{Coles2018etal}. 
In contrast, the latest quantum annealers support up to $2^{10}$ qubits \cite{DWAVE2000Q}\footnote{the amount of logical qubits 
which are available on this system is an order of magnitude lower since most qubits are used for the error correction}. 
Nevertheless, due to design and practical restrictions, quantum algorithms for the gates model such as Shor's prime number factorisation \cite{Shor1997} or Grover's search algorithms \cite{Grover1996} cannot be implemented on current quantum annealers. 
\noindent\textbf{Motivation and Contributions.} 
Considering recent successful applications of AQA in several fields of computational science \cite{Lloyd2013, Neukart2017, YLi2018}, we are motivated to investigate how useful AQA can be for computer vision and which problems can be potentially solved on the new hardware. 
The vast majority of available materials about quantum annealers are either oriented to physicists or lack technical details and clarity.
\textit{Our goal is to fill this gap, introduce the reader into the modern AQA and provide all notions and the background to understand, analyse, simulate and design quantum algorithms for computer vision which can potentially run on modern AQA, as well as interpret the results.} 
We consider correspondence problems on point sets which have various applications in computer vision. 
They consist in finding an optimal rigid transformation between inputs \cite{Horn88, Besl1992, Myronenko2010, Yang2016}. 
While transformation estimation assumes known matches, point set alignment is more general and targets, in addition, the recovery of correspondences. 
We consider two inputs, \textit{i.e.,} a fixed \textit{reference} point set and a \textit{template} undergoing a rigid transformation. 
\textit{Thus, our goal is to design a quantum approach for point set alignment which can potentially run on AQA and show that it offers advantages compared to the classical counterparts. }

Therefore, we adapt the recent progress in rigid point set alignment and formulate a globally multiply-linked energy functional which does not require any intermediate correspondence updates \cite{BHRGA2019}. 
In the \textit{gravitational approach} (GA) \cite{BHRGA2019},  the optimal alignment is achieved when the gravitational potential energy (GPE) of the system with two interacting particle swarms is locally minimal. 
Proceeding from GA, we build the weight matrix $\bfP$ for the associated QUBOP \eqref{eq:QUBOP} which is unalterably valid in the course of the %
optimisation. 
Along with that, we are targeting at a method which is implementable on classical hardware and can solve real-world problems, \textit{cf.}~Fig.~\ref{fig:DIFFERENT_2D_}. 
To summarise, the main \textbf{contributions} of this paper are: 
\begin{itemize}[leftmargin=*]%
    \vspace{-4pt} 
    \setlength{\itemsep}{7pt}
    \setlength{\parskip}{1.5pt}
    \item A self-contained and detailed introduction into modern quantum annealers for computer vision problems, including 
    notions from quantum physics and computing (Sec.~\ref{sec:preliminaries}),
    modern adiabatic quantum annealers (Sec.~\ref{sec:modern_AQC}) including D-WAVE (Sec.~\ref{sec:DWAVE}),  
    and previous and related works from quantum computing (Sec.~\ref{sec:related_work}). 
    \item The first quantum approach (QA) to transformation estimation (Sec.~\ref{sec:quantum_transformation_estimation}) and point set alignment (Sec.~\ref{sec:QPSR}) which can run on the upcoming quantum annealers (Sec.~\ref{ssec:complexity_of_QA}). 
    \item Experimental analysis of the proposed method in a simulated environment on several datasets  
    (Sec.~\ref{sec:EXPERIMENTS}). 
\end{itemize}

\begin{table}[]
    \centering
    \footnotesize
    \begin{tabular}{c|c}
        \textbf{quantum notion}               	& \textbf{classical counterpart}            \\ \hline 
        \textit{qubit (states $|0\rangle$ and $|1\rangle$)}& bit (states $0$ and $1$)             \\ 
        \textit{(time-dependent) Hamiltonian}  	& energy functional                         \\ 
        \textit{eigenstate}			& some energy state 			    \\ 
        \textit{ground state}                 	& globally optimal energy state             \\ 
        \textit{quantum system evolution}     	& optimisation process                      \\ 
        \textit{quantum annealing} \cite{Farhi2001} & simulated annealing \cite{Kirkpatrick1983} 
    \end{tabular} 
    \vspace{-4pt} 
    \caption{Quantum notions and their counterparts in computer vision. 
    }
    \vspace{-7pt}
    \label{tab:quantum_and_classic_notions} 
\end{table} 

\section{Preliminaries, Definitions and Notations}\label{sec:preliminaries} 

In this section, we introduce the reader into the basics of quantum computing. 
See Table~\ref{tab:quantum_and_classic_notions} for a lookup of notions specific to AQA 
which have counterparts and interpretation in the classical optimisation theory for computer vision. 
\noindent\textbf{Qubit.} 
Quantum computing encompasses tasks which can be performed on quantum-mechanical systems \cite{NielsenChuangQCQ2011}. 
Quantum \textit{superposition} and \textit{entanglement} are two forms of parallelism evidenced in quantum computers. 
A \textit{qubit} is a quantum-mechanical equivalent of a classical bit. 
A qubit $|\phi\rangle$ --- written in the \textit{Dirac} notation --- can be in the state $|0\rangle$, $|1\rangle$ or an arbitrary \textit{superposition of both states} denoted by $|\phi\rangle = \alpha |0\rangle + \beta |1\rangle$, where $\alpha$ and $\beta$ are the (generally, complex) probability amplitudes satisfying $|\alpha|^2 + |\beta|^2 = 1$. 
In quantum computing, the state $\frac{|0\rangle + |1\rangle}{\sqrt{2}}$ denoted by $|+\rangle$ is often used for initialisation of a qubit register. 
The state of a qubit remains hidden during the entire computation and reveals when measured. 
If qubits are \textit{entangled}, measuring one of them influences the measurement outcome of the other one \cite{Rieffel2000}. 
During the measurement, the qubit's state irreversibly collapses to one of the basis states $|0\rangle$ or $|1\rangle$. 
Efficient physical realisation of a qubit demand very low temperatures. 
Otherwise, thermal fluctuations will destroy it and lead to arbitrary changes of the measured qubit state. 
One possible physical implementation of a qubit is an electron which possesses a spin, \textit{i.e.,} its intrinsic magnetic moment \cite{NielsenChuangQCQ2011, Wesenberg2009}. 
The spin of an electron can be manipulated and brought to the state \textit{spin down}, \textit{spin up}, or a superposition of both. 
A concrete experimentally realised scheme that uses this property is represented by an atom of phosphorus $^{31}\text{P}$ embedded into a $^{28}\text{Si}$ silicon lattice attached to a transistor \cite{Kane1998,  Morello2010etal, Zwanenburg2013}. 
The nucleus of $^{31}\text{P}$ has a positive charge compensated by electrons. %
The bundle of electrons in the transistor is filled up to the energetic level between the energy of spin-down and spin-up state of $^{31}\text{P}$. 
To change a state of a $^{31}\text{P}$--$^{28}\text{Si}$ qubit, a microwave pulse of the frequency --- which is equal to the resonance frequency of the  atom --- is applied to it. 
The new state $|\phi\rangle$ depends on the duration of the exposure. 
A transistor is used to measure a state of the $^{31}\text{P}$--$^{28}\text{Si}$ qubit. 
If the extra electron of $^{31}\text{P}$ tunnels into the electron bundle, a positive charge is measured in the transistor indicating the spin-up state (\textit{e.g.,} $|1\rangle$). 
Fig.~\ref{fig:BLOCH_SPHERE}-(a) visualises a qubit with a so-called Bloch sphere. 
Every \textit{qubit} can be both in a superposition and entangled with other qubits. 
Thus, quantum superposition is the property that calculations are performed on all possible inputs simultaneously which can result in exponential parallelism in the number of qubits. 
When entangled, states of qubits cannot be described independently from each other. 
\definecolor{orange}{rgb}{1.,0.2,.0}
\begin{figure}[t!] 
\centering 
\includegraphics[width=1.0\linewidth]{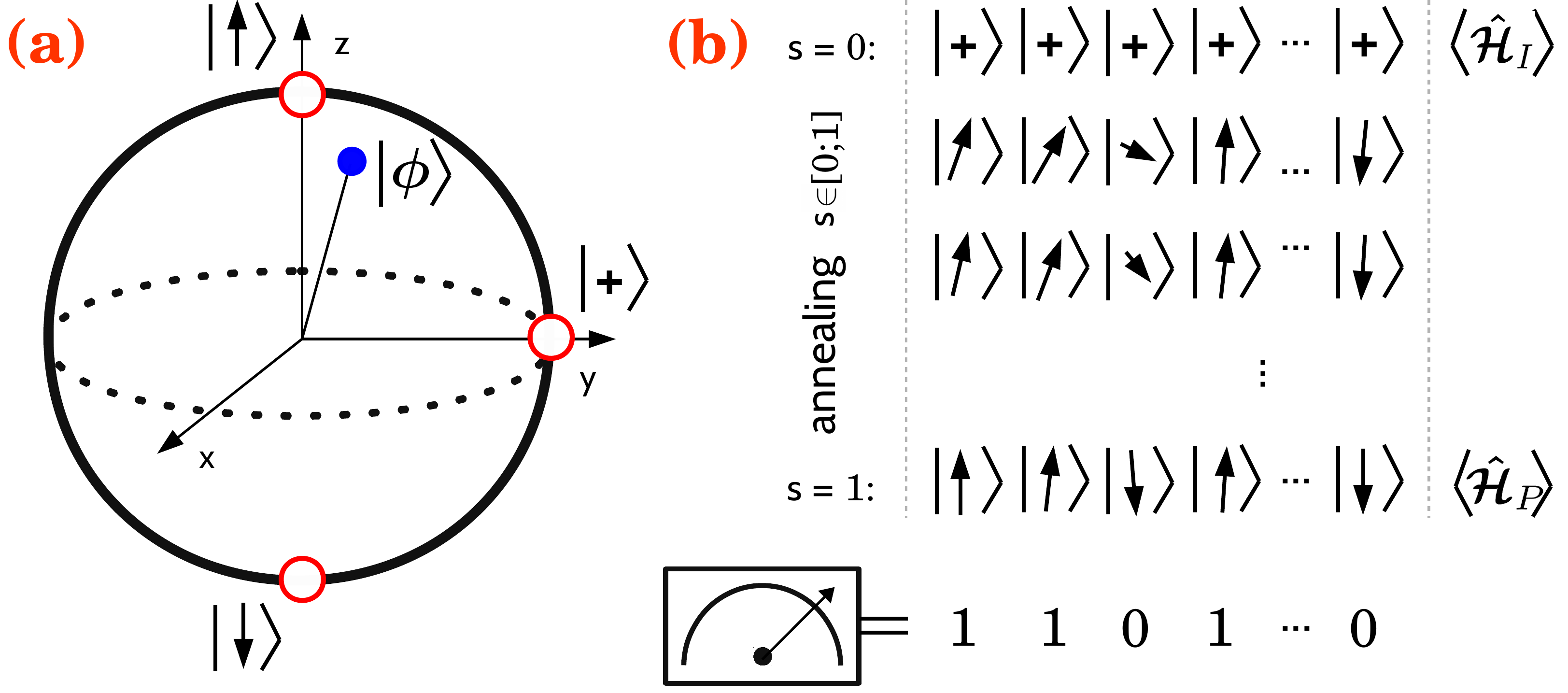} 
\caption{\textcolor{orange}{\textbf{(a)}}: Schematic depiction of a qubit with a Bloch sphere. 
Spin-up or $|1\rangle$ is located on the north pole, and spin down or $|0\rangle$ is located on the south pole. 
The state $\frac{|0\rangle + |1\rangle}{\sqrt{2}}$ with equal probability amplitudes to measure $|1\rangle$ and $|0\rangle$ values is geodetically equidistant to both poles. 
A point on the surface of the Bloch sphere corresponds to a valid pure state $|\phi\rangle = \alpha |0\rangle + \beta |1\rangle$. 
\textcolor{orange}{\textbf{(b)}}: Schematic visualisation of adiabatic quantum annealing (AQA). 
At the beginning, all qubits are initialised in the state $|+\rangle$. 
After the annealing is finished, the qubit states are measured and returned. 
After the measurement, the states of variables are classical. 
} 
\label{fig:BLOCH_SPHERE} 
\end{figure} 

\noindent\textbf{Schr\"odinger Equation. } 
In the universal or gates model, changes are expressed by a series of unitary transformations applied to qubits. 
This is a useful practical simplification, while the evolution of every quantum-mechanical system can be described more precisely 
by continuous Schr\"odinger equation, which in common notation reads: 
\begin{equation}\label{eq:SchroedingerEquation} 
  -i \, \frac{d}{dt} \, |\phi(t)\rangle = \Ham(t) \, |\phi(t)\rangle. 
\end{equation} 
For the simplicity, we denote here by $|\phi\rangle$ a state of $n$ qubits at time $t$, and $\Ham(t)$ is a Hamiltonian which 
is, in this case, a $2^n \times 2^n$ Hermitian matrix. 
Thus, a discrete time evolution of the quantum system is given by a unitary transformation. 
\\ 
\textbf{Hamiltonian.} Hamiltonian $\Ham$ is an energy operator of a system of $n$ qubits. 
It defines the energy spectrum of a system or, in our case, the space of all possible solutions. 
The \textit{ground state} of the system is its lowest energy eigenstate. 
Finding a ground state of a Hamiltonian is equivalent to finding an optimal solution to the problem. 
The expectation value of Hamiltonian $\langle\Ham\rangle$ provides an instantaneous energy of a given qubit configuration. 
In correspondence problems, $\langle\Ham\rangle$ is a quantitative characteristic of point set alignment. 
We denote by $\Delta(\Ham)$ the spectral gap of $\Ham$, \textit{i.e.,} the difference between the energies of the ground state and the second lowest eigenstate. 
The spectral gap influences the annealing rate and is considerable for algorithm design and evaluation in quantum annealing. 
\\ 
\textbf{Pauli Matrices.} An arbitrary Hamiltonian of a $n$-qubit-system can be expressed by a linear combination of tensor products of 
Pauli matrices denoted by: 
\vspace{-8pt} 
\begin{equation} 
   \small
   \boldsymbol{\sigma}^{x} = \begin{pmatrix} 0 & 1 \\ 1 & 0 \end{pmatrix}, \; \boldsymbol{\sigma}^{y} = \begin{pmatrix} 0 & -i \\ i & 0 \end{pmatrix},  \;\boldsymbol{\sigma}^{z} = \begin{pmatrix} 1 & 0 \\ 0 & -1 \end{pmatrix}. 
\end{equation} 
The Pauli matrices are $2 \times 2$ Hermitian and unitary. Together with the identity $\boldsymbol{\sigma}^{0} = \bfI_{2 \times 2}$, they %
form a basis for $\mathbb{C}^{2 \times 2}$. 
$\boldsymbol{\sigma}^{x}$ flips the probabilities to measure $|0\rangle$ and $|1\rangle$, whereas $\boldsymbol{\sigma}^{z} |0\rangle = |0\rangle$, and 
$\boldsymbol{\sigma}^{z} |1\rangle = -|1\rangle$. 
\\ 
\textbf{Pseudo-Boolean Functions}. A pseudo-boolean function is a real vector-valued function of $n$ boolean variables denoted by $\mathrm{x}$ of the form 
$\bfF(\mathrm{x}):\bfB^n \to \mathbb{R}^{\mathrm{M}}$, where $\mathrm{M}$ is the number of real-valued outputs. 
\\ 
\textbf{Quantum Annealing. } 
Quantum annealing is a heuristic combinatorial optimisation method for finding global optima which relies on quantum effects (superposition, entanglement and tunnelling) \cite{Finnila1994, Kadowaki1998}. In particular, it is used to find a ground state of an Ising Hamiltonian \cite{Ising1925, Peierls1936}, which encodes the target computational problem, see Fig.~\ref{fig:BLOCH_SPHERE}-(b). 

Quantum annealing is the quantum counterpart of simulated annealing \cite{Metropolis1953, Kirkpatrick1983}. 
Starting from the superposition state $[|+\rangle ]^{\otimes n}$ (this is a shorthand for $n$ qubits in the state $|+\rangle$, 
\textit{cf.}~\eqref{eq:initial_state}), the system evolves according to~\eqref{eq:SchroedingerEquation} under an external time-dependent magnetic field (a transverse field). 
When the external field is faded away, the system reaches the ground state of an Ising model \cite{Ising1925}. %
According to~\eqref{th:adiabatic_theorem}, if an external magnetic field is changing gradually enough, the system remains near the ground state 
with high probability throughout the optimisation. %
Quantum annealing systems taking advantage of~\eqref{th:adiabatic_theorem} are called %
\textit{adiabatic quantum computers} (AQC). 
QUBOP is the most common problem form which can be mapped to current realisations of AQC. %

\section{Modern Adiabatic Quantum Computation}\label{sec:modern_AQC} 

Adiabatic quantum computation is a form of quantum annealing which relies on the adiabatic theorem of quantum mechanics~\eqref{th:adiabatic_theorem} \cite{Born1928}. %
Starting from a ground state of an initial default Hamiltonian $\Ham_I$, an AQC system adiabatically evolves into the ground state of a problem Hamiltonian $\Ham_P$ which encodes a solution to a problem \cite{Farhi2001}. 
In the case of \textit{adiabatic quantum annealing} (AQA), the problem Hamiltonian $\Ham_P$ is given by the Ising model \cite{Ising1925}: 
\begin{equation}\label{eq:Ising_Model_Hamiltonian} 
  \Ham_P = \sum_{j \in V} \, h_{j} \boldsymbol{\sigma}_j^{z} \;\; + \sum_{(j,\,k) \in E_P} J_{j, k} \, \boldsymbol{\sigma}_j^{z} \otimes \boldsymbol{\sigma}_k^{z}, 
\end{equation} 
with the Kronecker product $\otimes$, 
$h_{j}$ denoting exterior local magnetic fields and $J_{i, j}$ standing for the pairwise connections between the particles. 
$V$ is a set of particles, and $E_P$ is a set of edges (intra-particle links) of the graph. %
Eq.~\eqref{eq:Ising_Model_Hamiltonian} is written in a notation common in physics. %
The first term of \eqref{eq:Ising_Model_Hamiltonian} on the right side in the explicit notation reads 
\begin{equation}\label{eq:initial_hamiltonian_explicit} 
  \footnotesize 
  \Ham_P^{j \in V} = \begin{pmatrix} 
		\begin{bmatrix} 
		\begin{bmatrix} 
		  \boldsymbol{\sigma}^{z} \otimes \bfI \otimes \hdots \otimes \bfI 
		\end{bmatrix} \\
		\begin{bmatrix}
		  \bfI \otimes \boldsymbol{\sigma}^{z} \otimes \hdots \otimes \bfI 
		\end{bmatrix} \\
		\ddots	      \\
		\begin{bmatrix}
		  \bfI \otimes \bfI \otimes \hdots \otimes \boldsymbol{\sigma}^{z} 
		\end{bmatrix}
		\end{bmatrix}^\mathsf{T}  
		\end{pmatrix}_{2^n \times n 2^n } %
		\begin{bmatrix} 
		  h_1 \,\bfI_{2^n \times 2^n} 	\\ 
		  h_2 \,\bfI_{2^n \times 2^n} 	\\ 
		  \vdots 		    	\\ 
		  h_n \, \bfI_{2^n \times 2^n} 	\\ 
		\end{bmatrix}_{n 2^n \times 2^n},  
\end{equation} 
where $\bfI$ without a subscript is a $2 \times 2$ identity matrix. 
The second term $\Ham_P^{(j,\,k) \in E_P}$ of \eqref{eq:Ising_Model_Hamiltonian} can be expressed in a similar manner, involving 
pairs of  $\boldsymbol{\sigma}^{z}$ in the tensor product depending on the connectivity of the lattice.
Theoretically, each particle can interact with any other particle from the whole set of qubits. %
In practice, the couplings are restricted to local neighbourhoods (see Sec.~\ref{sec:DWAVE}). %
Thus, \eqref{eq:Ising_Model_Hamiltonian} describes a system of $N$ interacting spin-\textonehalf~particles under the influence of distributed magnetic forces, and in the expanded form, $\Ham_P$ is a $2^n \times 2^n$ matrix. 
Finding a ground state of an Ising model is an $\mathcal{NP}$-hard problem \cite{Barahona1982}. 
In the ground state, the spin configuration of all particles which minimises Ising energy $\bfE_{\text{Ising}}$ is given by: 
\begin{equation} 
  \bfE_{\text{Ising}} = \sum_i h_i s_i + \sum_{i, j} J_{i, j} s_i s_j, 
\end{equation} 
where $s_i \in \{1, -1\}$ denotes two possible spin measurement outcomes of a spin-\textonehalf~particle.

\subsection{Quantum System Evolution}\label{sec:system_evolution}

Solving $\mathcal{NP}$-hard problems such as QUBOP on a classical computer requires exponential time in the size of the input. 
The main idea of the AQC is that a QUBOP \eqref{eq:QUBOP} can be mapped to the Ising model \eqref{eq:Ising_Model_Hamiltonian} 
and optimised by allowing the system to evolve according to the adiabatic principle \eqref{th:adiabatic_theorem}. 
Once annealing is finished, the qubit register will represent the solution to the programmed problem with a high probability \cite{Farhi2001} (\textit{cf.}~Appendix~\ref{app:appendix} on the annealing rate criterion). 
The initial Hamiltonian of the system is always initialised in the state 
\begin{equation}\label{eq:initial_Hamiltonian} 
  \Ham_I = - \sum_{j \in V} B_x \boldsymbol{\sigma}_j^{x}, %
\end{equation} 
where $B_x > 0$ stands for a magnetic field pointing in the $x$ direction. 
The ground state of \eqref{eq:initial_Hamiltonian} is a symmetrised superposition with equal normalised probability amplitudes 
for the states $|0\rangle$ and $|1\rangle$ for all qubits, \textit{i.e.,} 
\begin{equation}\label{eq:initial_state} 
  [|+\rangle ]^{\otimes n} = \bigg( \frac{|0\rangle + |1\rangle}{\sqrt{2^{n}}} \bigg)^{\otimes n}. 
\end{equation} 
This initial state \eqref{eq:initial_state} is comparably easy to construct by radiating a microwave of the same duration and wavelength to all qubits. 
In mathematical terms, \eqref{eq:initial_state} is obtained by applying a Hadamard transform {\footnotesize $H = \frac{1}{\sqrt{2}}\begin{bmatrix}
                                                                                                         1 & 1 \\ 1 & -1 
                                                                                                        \end{bmatrix}$} to $n$ $|0\rangle$ qubits.

The lowest energy $E_{\text{GS}}=-n B_x$ of \eqref{eq:initial_Hamiltonian} is achieved when all qubits in the system point in the anti-parallel direction of the magnetic field, so that $\boldsymbol{\sigma}_j^{x}|s_j \rangle =|s_j \rangle$. 
During AQC, the initial Hamiltonian $\Ham_I$ is evolving into the problem Hamiltonian $\Ham_P$, with a high probability of reaching 
the ground state of $\Ham_P$ \cite{Farhi2001}. 
The interpolation between the Hamiltonians can be written as 
\begin{equation} 
  \Ham = [1 - s] \, \Ham_I + s \, \Ham_P, 
\end{equation} 
with $s \in [0;1]$ being the time in relative units from the start of annealing at $s = 0$ until reaching the ground state of $\Ham_P$ at $s = 1$. 
The problem Hamiltonian and the final state of the system depend on the objective function $f(x)$ or the matrix of weights between the qubits $\bfP$ in \eqref{eq:QUBOP}. 
After the annealing is accomplished, the state of each qubit is measured, and the result corresponds to the solution of the programmed problem with a high probability. 
At this stage, the states of all binary variables are classical, and not quantum anymore. 
To remain in the ground state during the system evolution, the annealing rate has to be carefully chosen. 
The condition of adiabaticity \eqref{th:adiabatic_theorem} is derived from the time-dependent perturbation theory of quantum systems. 
It is achieved when the average energy pumped into the system per time interval $T$ is smaller than the minimal energy difference between the ground state and the first excited state. 
This statement was quantified in \cite{Amin2009} which generalises the original adiabatic theorem \cite{Born1928} for periodic driving, 
see Appendix~\ref{app:appendix} for further details. 

\subsection{Quantum Annealer D-WAVE}\label{sec:DWAVE}

D-WAVE relies on the adiabatic criterion in its specified form and currently supports up to ${\approx}2000$ qubits \cite{DWAVE_QPU}. 
It reflects the state of the art in physical realisation of quantum processors. 
It is relatively inexpensive to bring the system in the superposition state, and every computation on D-WAVE starts with the problem-independent $\Ham_I$ \eqref{eq:initial_Hamiltonian}. 
Qubits can interact with a restricted number of other qubits, and it is possible to define qubit equality and entanglement constraints \cite{DWAVE_QPU}. 
Possible interactions can be seen from the \textit{chimera} graph which schematically depicts the layout of the quantum processor \cite{DWAVE_QPU, Boothby2016}. 
At the same time, the physically realised connectivity can model QUBOP with arbitrary connectivities through an internal conversion \cite{Boothby2016}. 
The drawback is that in the worst case, a quadratic increase in the number of variables is required. %
A fully connected graph of layers with $N$ qubits would require $N^2$ qubits for processing. %
Some QUBOP cannot be mapped to the chimera graph, and some problems can be mapped in multiple ways \cite{Parekh2016}. 

\section{Previous and Related Work}\label{sec:related_work}

\noindent\textbf{Universal Quantum Computers.} 
The paradigm of the universal quantum computer originates in the attempts to gain control over %
individual quantum systems in the early eighties \cite{Shor2001, NielsenChuangQCQ2011}. Later, extending the control to multiple quantum systems 
has attracted the interest of physicists, promising to facilitate discoveries in quantum physics \cite{NielsenChuangQCQ2011}. %
By that time, it was noticed that simulating a quantum-mechanical system on a classical computer requires exponential time 
in the number of simulated elements \cite{Manin1980, Feynman1982}. 
``Can you do it\footnote{to simulate quantum-mechanical effects} with a new kind of computer -- a quantum computer?'' \cite{Feynman1982} is 
a famous quote by R.~Feynman which has triggered research on quantum computers in the subsequent years. %
The so-called \textit{no-cloning theorem} \cite{Park1970, Wootters1982} belongs to the first discoveries strongly influenced quantum information theory and quantum computations. 
Nowadays, quantum computers can be used not only to fulfil their primary goal, \textit{i.e.,} to simulate quantum-mechanical systems 
for different branches of science, but also to solve other computational problems --- such as balanced function decision problem \cite{DeutschJozsa1992RSP}, 
quantum Turing machines for complexity analysis \cite{BernsteinVazirani1993}, prime number factorisation and discrete logarithms \cite{Shor1997}, 
database search \cite{Grover1996}, graph matching \cite{AmbainisSpalek2006}, data classification \cite{Rebentrost2013} and principal component analysis \cite{Lloyd2013} --- faster than on classical machines. 
The related field of quantum communication and quantum key distribution has already found broad practical use nowadays 
\cite{BennettBrassard1984, Bennett_1992, SchmittManderbach2007}. 

\noindent\textbf{Classical Methods using Quantum Analogies.} 
Quantum-mechanical effects inspired multiple techniques for conventional computers including variants of genetic and evolutionary algorithms \cite{Han2000, HanKim2002}, non-rigid mesh analysis \cite{Aubry2011} and image segmentation \cite{Aytekin2014}, among others. 
\noindent\textbf{Quantum Annealers in Computer Vision.} 
Only a few theoretical results and applications of AQC to image processing, machine learning and computer vision are known. 
Neven~\textit{et al.}~\cite{Neven2008arXiv} have shown how image recognition can be formulated as QUBOP. 
Image classification on $12 \times 12$ images with AQC was addressed in \cite{Nguyen2019arXiv}. 
The approach of O'Malley~\textit{et al.} can learn facial features and reproduce facial image collections \cite{OMalley2018}. 
Boyda~\textit{et al.}~\cite{Boyda2017} propose an AQC method to detect areas with trees from aerial images. 
Several methods target classification, dimensionality reduction and training of deep neural networks \cite{Neven2012, Khoshaman2018, Adachi2015arXiv}. 
Not all theoretical findings of these works are possible to test on the real AQC hardware yet. 
Nonetheless, we believe that it is essential to explore the theory and highlight the advantages of the upcoming hardware for computer vision tasks. 

\section{Quantum Transformation Estimation}\label{sec:quantum_transformation_estimation} 

In this section, we introduce our QA to transformation estimation. 
The inputs are a reference point set $[\bfx_n] \in \bfX \in \mathbb{R}^{D \times N}$ and a template point set $[\bfy_n] \in \bfY \in \mathbb{R}^{D \times N}$, $n \in \{1, \hdots, N\}$. 
$N$ is the number of points in both point sets and $D$ is the dimensionality of the points. 
We assume that translation is resolved, the centroids of the point sets coincide, and points are in correspondence. 

\subsection{Transformation Estimation in 2D}\label{sec:2D_transformation_estimation}

To obtain an advantage in solving transformation estimation on a quantum annealer we should avoid uniform sampling of rotations applied to $\bfY$. 
Elements of the rotation group are non-commutative, and it is not possible to formulate multiplication of basis rotations as QUBOP. 
Instead, we propose to represent the transformation matrix as a linear combination of basis elements. 
Recall that for any rotation matrix, $\bfR^{-1} = \bfR^{\mathsf{T}}$. 
Rotation in 2D consists of four elements, \textit{i.e.,} {\small$\bfR = \begin{pmatrix} r_{1,2} & r_{2,2} \\ r_{2,1} & r_{2,2} \end{pmatrix}$}. 
Additively, we can create a basis for all possible values of $\bfR$ and encode the influence of the additive elements as binary variables. 
Consider instead the power series of $\bfR$ in 2D. 
Every such matrix has a corresponding skew-symmetric matrix of the form %
\begin{equation} 
  \bfS = \theta \, \bfM, \;\;\; \bfM =  \begin{bmatrix} 
		    0 & -1 \\ 
		    1 & 0 
		    \end{bmatrix}, 
\end{equation} 
with a real number $\theta$. 
According to the Cayley-Hamilton theorem, $\bfS^2 + \theta^2 \bfI = 0$ which leads to the following exponential map for $\bfR$ with power series: 
\begin{equation}\label{eq:exp_map} 
\begin{aligned} 
  & \bfR = \exp (\bfS) = \\ 
  & \cos(\theta) \, \bfI + \bigg( \frac{\sin(\theta)}{\theta} \bigg) \, \bfS = \cos(\theta) \, \bfI + \sin(\theta) \, \bfM.   
\end{aligned} 
\end{equation} 
From \eqref{eq:exp_map} we see that $\bfR$ is composed of an identity weighted by $\cos(\theta)$ and $\bfM$ weighted by $\sin(\theta)$. %
If the basis would resemble additive elements $\bfI$ and $\bfM$ of the exponential map, we can stronger constrain the resulting $\bfR$. %
We see that $r_{1,1}$ is entangled with  $r_{2,2}$, and $r_{1, 2}$ is entangled with $r_{2,1}$. 
Eventually, we need fewer basis elements, the optimisation will finish faster and the method can be also implemented and tested on a classical computer. 
Thus, our basis $\bfQ = \{\bfQ_k\}$ for $\bfR$ is a compound of $K = 20$ elements: 
\begin{equation}\label{basis_2D} 
\begin{aligned}
  \hspace{-7pt} \big\{\bfQ_{k} = \omega \, \bfC \in \mathbb{R}^{2 \times 2}, 
   &\forall \omega \in \{0.5, 0.2, 0.1, 0.1, 0.05\},  \\ %
  &\forall \bfC \in \{\bfI, \bfM, -\bfI, -\bfM\} \big\}. 
\end{aligned} 
\end{equation} 
Since we want to find $\bfR$ which minimises the distances between the corresponding points $(\bfx_n, \bfy_n)$, we multiply each template point with a \textit{negative} sign $-\bfy_n$ with each basis element $\bfQ_k$ and stack the result into  $\boldsymbol{\Phi}$: %
\begin{equation}\label{eq:PHI} 
   \small 
   \boldsymbol{\Phi} = \begin{bmatrix} 
                         \bfx_1^\mathsf{T} 		& \bfx_2^\mathsf{T} 		& \hdots 	& \bfx_N^\mathsf{T} 			\\ 
                         -[\bfQ_1 \bfy_1 ]^\mathsf{T} 	& -[\bfQ_1 \bfy_2]^\mathsf{T}	& \hdots 	& -[\bfQ_1 \bfy_N ]^\mathsf{T}  	\\ 
                         -[\bfQ_2 \bfy_1 ]^\mathsf{T} 	& -[\bfQ_2 \bfy_2 ]^\mathsf{T}	& \hdots 	& -[\bfQ_2 \bfy_N ]^\mathsf{T}  	\\ 
                         \vdots				& \vdots 			& \ddots 	& \vdots 				\\ 
                         -[\bfQ_K \bfy_1 ]^\mathsf{T} 	& -[\bfQ_K \bfy_2 ]^\mathsf{T}	& \hdots 	& -[\bfQ_K \bfy_N ]^\mathsf{T}  	\\ 
                       \end{bmatrix}. %
\end{equation} 
Next, we set the weight matrix in \eqref{eq:QUBOP} as 
\begin{equation}\label{eq:PHI_PHI_T} 
 \bfP = \boldsymbol{\Phi} \boldsymbol{\Phi}^{\mathsf{T}}, 
\end{equation} 
and the final QUBOP reads 
\begin{equation}\label{eq:xtPx} 
  \arg \min_{\bfq \in \bfB^{21}} \bfq^\mathsf{T} \boldsymbol{\Phi} \boldsymbol{\Phi}^{\mathsf{T}} \bfq. 
\end{equation} 
In total, $21$ qubits are required to resolve the transformation on AQC in 2D, with the first qubit of $\bfq$ being fixed to $|1\rangle$. 
After solving \eqref{eq:xtPx} with quantum annealing and measuring $\bfq$, we obtain a classical bitstring $\hat{\bfq}$. 
The resulting (perhaps approximate) $\bfR$ is then obtained by \textit{unembedding} as 
\begin{equation}\label{eq:unembedding} 
  \bfR = \sum_{k=1}^K \hat{\bfq}_{k+1} \bfQ_k. 
\end{equation} 

\subsection{Transformation Estimation in 3D}\label{sec:3D_transformation_estimation}

In 3D, a skew-symmetric matrix can be represented as 
\begin{equation} 
  \bfS = \theta \, \bfM, \; \bfM = \begin{bmatrix} 
                                        m_{1,1} & m_{1,2} & m_{1,3} \\ 
                                        m_{2,1} & m_{2,2} & m_{2,3} \\ 
                                        m_{3,1} & m_{3,2} & m_{3,3}
                                       \end{bmatrix} = 
                                       \begin{bmatrix} 
                                        0 	& a 	& b \\ 
                                        -a	& 0	& c \\ 
                                        -b	& -c	& 0 
                                       \end{bmatrix}, 
\end{equation} 
where $\theta$, $a$, $b$ and $c$ are real numbers, and $a^2 + b^2 + c^2 = 1$. 
In the 3D case, the Cayley-Hamilton theorem states that $-\bfS^3 - \theta^2 \bfS = 0$. 
The exponential map for $\bfR$ in 3D with power series reads 
\begin{equation}\label{eq:exp_map_3D} 
  \begin{aligned} 
   \bfR = \exp (\bfS) = &\bfI + \bigg( \frac{\sin \theta}{\theta} \bigg) \, \bfS + \bigg( \frac{1 - \cos \theta}{\theta^2} \bigg) \, \bfS^2 = \\ 
  & \bfI + \sin \theta \, \bfM + (1 - \cos \theta) \, \bfM^2. 
  \end{aligned} 
\end{equation} 
Next, $\bfM$ can be decomposed as follows: 
\begin{equation}
\small
\begin{aligned} 
  \bfM =  \underbrace{a \begin{bmatrix}
	      0  & 1 & 0 \\
	      -1 & 0 & 0 \\ 
	      0  & 0 & 0 
           \end{bmatrix}}_{\text{\normalsize $\bfM_a$}} + 
           \underbrace{b \begin{bmatrix}
	      0  & 0 & 1 \\
	      0 & 0 & 0 \\ 
	      -1  & 0 & 0 
           \end{bmatrix}}_{\text{\normalsize $\bfM_b$}} + 
           \underbrace{c \begin{bmatrix}
	      0  & 0 & 0 \\
	      0 & 0 & 1 \\ 
	      0  & -1 & 0 
           \end{bmatrix}}_{\text{\normalsize $\bfM_c$}}. 
\end{aligned} 
\end{equation} 
Regarding $\bfM$, we see that 
\begin{itemize} 
 \item $\{m_{1,2}; m_{2,1}\}$, $\{m_{1,3}; m_{3,1}\}$ and $\{m_{2,3}; m_{3,2}\}$ are mutually dependent or entangled, 
 \item $m_{i,j} \in [-1; 1]$, 
 \item $\bfM = -\bfM^{\mathsf{T}}$, $\bfM_a = - \bfM_a^{\mathsf{T}}$, $\bfM_b = - \bfM_b^{\mathsf{T}}$ and $\bfM_c = - \bfM_c^{\mathsf{T}}$, \textit{i.e.,} they are anti-symmetric, and  
 \item ${\small \bfM^2 = \begin{bmatrix}	 
                   v_1^- & d     & e \\ 
                   d     & v_2^- & f \\ 
                   e     & f     & v_3^- 
                 \end{bmatrix}}$
is symmetric negative semi-definite, with $\{v_1^-, v_2^-, v_3^-\} \in \mathbb{R}^-$, and $\{d, e, f\} \in \mathbb{R}$. 
\end{itemize} 
The basis for rotation in 3D is comprised of the identity matrix $\bfI$, $\bfM_a$, $\bfM_b$, $\bfM_c$ as well as the basis for $\bfM^2$: 
\begin{equation}
\small
  \bfM_d = \begin{bmatrix} 
             0 & 1 & 0 \\ 
             1 & 0 & 0 \\ 
             0 & 0 & 0 
           \end{bmatrix}, 
  \bfM_e = \begin{bmatrix} 
             0 & 0 & 1 \\ 
             0 & 0 & 0 \\ 
             1 & 0 & 0 
           \end{bmatrix}, 
  \bfM_f = \begin{bmatrix} 
             0 & 0 & 0 \\ 
             0 & 0 & 1 \\ 
             0 & 1 & 0 
           \end{bmatrix}. 
\end{equation} 
Thus, our basis $\bfQ^{3D} = \{\bfQ^{3D}_k\}$ for $\bfR$ in 3D is a compound of $K = 80$ elements: 
\begin{equation}\label{basis_3D} 
\hspace{-7pt} 
\begin{aligned} 
  & \big\{\bfQ^{3D}_{k} = \omega \, \bfC^{3D} \in \mathbb{R}^{3 \times 3},  
  \forall \omega \in \{0.5, 0.2, 0.1, 0.1, 0.05\},  \\ %
  & \forall \bfC^{3D} \in \{ \bfI, -\bfI, \; \bfM_a, -\bfM_a, \bfM_b, -\bfM_b, \bfM_c, -\bfM_c, \\ 
  & \;\;\;\;\;\;\;\;\;\;\;  \bfM_d, -\bfM_d, \bfM_e, -\bfM_e, \bfM_f, -\bfM_f\} \big\}. 
\end{aligned} 
\end{equation} 
The final QUBOP and the unembedding (\textit{i.e.,} decoding the solution to QUBOP) after quantum annealing for the 3D case are obtained similarly to  \eqref{eq:PHI}--\eqref{eq:unembedding} 
with $\bfq \in \bfB^{81}$ ($\bfq_0$ remains fixed to $|1\rangle$ and $\bfQ_k$ are  replaced by $\bfQ_k^{3D}$ in \eqref{eq:unembedding}). 

\section{Quantum Point Set Registration}\label{sec:QPSR} 

In point set registration, the input point sets are of different cardinalities, and correspondences between points are, generally, not known, \textit{i.e.,} $[\bfx_n] \in \bfX \in \mathbb{R}^{D \times N}$ and $[\bfy_m] \in \bfY \in \mathbb{R}^{D \times M}$, $m \in \{1, \hdots, M\}$. %
$N$ and $M$ are the numbers of points in the reference and template, respectively, while $D$ is the point dimensionality. 
The objective of point set alignment is to recover rotation $\bfR$ ($\bfR^{-1} = \bfR^{\mathsf{T}}$, $\operatorname{det}(\bfR) = 1$) and translation $\bft$ aligning $\bfY$ to $\bfX$. 
We assume that the translation is resolved in the pre-processing step by bringing the point set centroids into coincidence. 

Point set alignment can be alternatingly solved on AQC by finding some point matches and estimating the transformation with the given correspondences in the ICP fashion \cite{Besl1992}. 
This would result in a sequence of QUBOP of the form \eqref{eq:xtPx}. 
To express alignment as a single QUBOP, we have to find an energy functional which is correspondence-free and which, when minimised in one shot on AQC, would result in an optimal alignment. 
The desired form of the energy functional has been recently shown in the literature \cite{BHRGA2019}. 
\subsection{Particle Dynamics Based Alignment}\label{ssec:particle_dynamics_based_alignment} 

Barnes-Hut Rigid Gravitational Approach (BHRGA) \cite{BHRGA2019} is a recent point set alignment method with a single energy functional which remains unchanged during the entire optimisation. 
BHRGA is a globally multiply-linked approach, \textit{i.e.,} all $\bfy_m$ interact with all $\bfx_n$. %
In \cite{BHRGA2019},  point sets are aligned by minimising the mutual \textit{gravitational potential energy} (GPE) $\bfE$ of the corresponding system of particles in the force field induced by $\bfX$: 
\begin{equation}\label{eq:our_final_GPE} 
  \bfE(\bfR, \bft) = \sum_m \sum_n \mu_{\bfy_m} \, \mu_{\bfx_n} \norm{\bfR \, \bfy_m + \bft - \bfx_n}_2, 
  \vspace{-5pt} 
\end{equation} 
where $\mu_{\bfy_m}$ and $\mu_{\bfx_n}$ denote masses of $\bfy_m$ and $\bfx_n$,  respectively. 
With no imposed boundary conditions, particles are initialised with unit masses. 
In \cite{BHRGA2019}, \eqref{eq:our_final_GPE} is optimised with the Levenberg-Marquardt algorithm \cite{Levenberg_44, Marquardt_1963}, and the optimum is achieved when the system's GPE is locally minimal. 
Without acceleration by a $2^D$-tree, the method has quadratic complexity and \eqref{eq:our_final_GPE} involves all possible interactions between the template and reference points. 

We can now derive a QUBOP in the similar fashion as in Sec.~\ref{sec:quantum_transformation_estimation} for the transformation estimation. 
Note, however, that the bases \eqref{basis_2D} and \eqref{basis_3D} allow for affine transformations and scaling. 
Thus, implicitly, we would optimise %
\begin{equation}\label{eq:our_final_GPE_scaling} 
  \hspace{-4pt}\bfE(\bfR, \bft, s) = \sum_m \sum_n \mu_{\bfy_m} \, \mu_{\bfx_n} \norm{\bfR \, \bfy_m s + \bft - \bfx_n}_2, 
\end{equation} 
where the scalar $s$ is the scaling of the template. 
As proven in \cite{proof_singularity_2019}, \textit{allowing for scale in globally multiply-linked point set alignment results in the shrinkage of the template to a single point with a very high probability}. 
To remedy the problem, either prior correspondences can be used, or point interactions can be restricted to local vicinities \cite{proof_singularity_2019}. 
In our QA, we opt for the second solution which allows to use the rotational bases \eqref{basis_2D} and \eqref{basis_3D} elaborated in Sec.~\ref{sec:quantum_transformation_estimation}. 
Eventually, the $\boldsymbol{\Phi} \in \mathbb{R}^{(K+1) \times (D)(L(1) + L(2) + \hdots + L(N))}$ matrix encoding point  interactions for point set alignment reads 
\begin{equation} 
    \boldsymbol{\Phi} = \big[ \boldsymbol{\Phi}_1 \boldsymbol{\Phi}_2 \hdots  \boldsymbol{\Phi}_N  \big],  
\end{equation} 
with $\boldsymbol{\Phi}_n$, $n \in \{1, \hdots, N\}$, of the form 
\begin{equation}\label{eq:Phi_PSR} 
    \small 
    \begin{bmatrix} 
			\bfx_n^\mathsf{T}   &  \bfx_n^\mathsf{T} 	& \hdots			& \bfx_n^\mathsf{T} 		\\ 
			-[\bfQ_1 \bfy_1^n ]^\mathsf{T} & -[\bfQ_1 \bfy_2^n ]^\mathsf{T}	& \hdots			& -[\bfQ_1 \bfy_{L(n)}^n ]^\mathsf{T}  \\ 
			-[\bfQ_2 \bfy_1^n ]^\mathsf{T} 	& -[\bfQ_2 \bfy_2^n ]^\mathsf{T} 	&\hdots			& -[\bfQ_2 \bfy_{L(n)}^n ]^\mathsf{T}  \\ 
                        \vdots		&  \vdots	& \ddots 			& \vdots 			\\ 
                        -[\bfQ_K \bfy_1^n ]^\mathsf{T} 	& -[\bfQ_K \bfy_2^n ]^\mathsf{T} 	& \hdots 			& -[\bfQ_K \bfy_{L(n)}^n ]^\mathsf{T}   
                      \end{bmatrix}, 
\end{equation} 
with $\bfQ_k$ being as in \eqref{basis_2D} or \eqref{basis_3D} for the 2D and 3D case, respectively. 
$\boldsymbol{\Phi}_1$, $\boldsymbol{\Phi}_2$ and $\boldsymbol{\Phi}_N$ encode point interactions between every $\bfx_n$ and corresponding $L(n) \ll M$ points of the template denoted by superscripted $\{\bfy_1^n, \bfy_2^n, \hdots, \bfy_{L(n)}^n\}$. 
%
%
Note that the latter build $N$ subsets of $\{\bfy_1,  \bfy_2, \hdots, \bfy_M\}$ of different cardinalities $L(n)$, $\bar{L}$ on average. 
If different $\bfx_n$ interact with the same $\bfy_m$, the corresponding subcolumns $\begin{bmatrix} [\bfQ_1 \bfy_m ]^{\mathsf{T}} [\bfQ_2 \bfy_m ]^{\mathsf{T}} \hdots [\bfQ_K \bfy_m]^{\mathsf{T}} \end{bmatrix}^{\mathsf{T}}$ of $\boldsymbol{\Phi}_n$ can be computed only once and reused. 
The final QUBOP for point set alignment with $\boldsymbol{\Phi}_n$ as in \eqref{eq:Phi_PSR} reads
\vspace{-1pt} 
\begin{equation}\label{QUBOP_PSR} 
  \arg \min_{\bfq \in \bfB^{K+1}} \bfq^\mathsf{T} \boldsymbol{\Phi} \boldsymbol{\Phi}^{\mathsf{T}} \bfq. 
  \vspace{-2pt} 
\end{equation} 
In total, $K + 1= 21$ and $K + 1= 81$ qubits are required to align point sets on AQC in the 2D and 3D case, respectively. 
Both transformation estimation and point set alignment need the same number of qubits in the same dimensions, and the difference lies in the complexity 
to construct $\bfP$ (see Sec.~\ref{ssec:complexity_of_QA}). %
Note that if the same template has to be aligned to multiple references, the corresponding $\boldsymbol{\Phi}$ can be obtained by reusing 
$\begin{bmatrix} [\bfQ_1 \bfy_m]^{\mathsf{T}} [\bfQ_2 \bfy_m ]^{\mathsf{T}} \hdots [\bfQ_K \bfy_m ]^{\mathsf{T}} \end{bmatrix}^{\mathsf{T}}$ (which has to be computed only once). 
The first qubit of $\bfq$ has to be fixed to $|1\rangle$, since the first element of every column contains a reference point which has to be active during the entire optimisation. 
The unembedding is performed similarly to the case of transformation estimation, see Sec.~\ref{sec:quantum_transformation_estimation}. 

\subsection{Complexity to Prepare {\large $\bfP = \boldsymbol{\Phi}  \boldsymbol{\Phi}^{\mathsf{T}}$}}\label{ssec:complexity_of_QA} 

To prepare $\boldsymbol{\Phi}$, $\mathcal{O}(K D N \xi)$ and 
$\mathcal{O}(K D N \bar{L} \xi)$ operations are required for the transformation estimation and point set alignment, respectively. $\xi$ denotes the number of operations for multiplying $\bfy_m$ with one element of the additive basis $\bfQ_k$. 
To obtain the final $\bfP$, we need to transpose $\boldsymbol{\Phi}$ and multiply $\boldsymbol{\Phi}$ with $\boldsymbol{\Phi}^{\mathsf{T}}$ which, in the worst case, takes $\mathcal{O}(K^2 DN)$ operations for the transformation estimation and $\mathcal{O}(K^2 D N \bar{L})$ operations for the point set alignment. 
There are also slightly faster algorithms for matrix multiplication compared to the na\"{i}ve way \cite{CoppersmithWinograd1990}. 

\begin{table}[t] 
  \center 
  \footnotesize 
    \begin{center} 
      \begin{tabular}{|c||c|c|c|c|c|c|} \cline{2-7} %
	\multicolumn{1}{c|}{}	& 	\multirow{2}{*}{\footnotesize \textbf{\textit{TE}}}	& 	\multicolumn{5}{c|}{\footnotesize $\boldsymbol{K}$}			\\\cline{3-7} 
	\multicolumn{1}{c|}{}	& 			&	$\boldsymbol{10}$ 	& 	$\boldsymbol{20}$	& 	$\boldsymbol{30}$ 	& $\boldsymbol{40}$ & $\boldsymbol{50}$	\\\hline %
	$e_{2D}$ 		& 	\cellcolor{green!25}$0.023$ 	&	\cellcolor{green!25}$0.026$			& 	\cellcolor{green!25}$0.041$			&	\cellcolor{green!45}$0.078$			& \cellcolor{blue!25}$0.17$	& \cellcolor{blue!50}$0.3$	\\\hline %
	$\sigma_{2D}$ 		& 	\cellcolor{green!25}$0.012$ 	&	\cellcolor{green!25}$0.013$			& 	\cellcolor{green!25}$0.012$			&	\cellcolor{green!45}$0.012$			& \cellcolor{blue!25}$0.012$	& \cellcolor{blue!50}$0.013$	\\\hline %
	$e_{\bfR}$		& 	\cellcolor{green!25}$0.058$ 	&	\cellcolor{green!25}$0.062$			& 	\cellcolor{green!45}$0.083$			&	\cellcolor{blue!25}$0.22$			& \cellcolor{blue!50}$0.47$ 	& \cellcolor{blue!70}$0.764$	\\\hline %
	$\sigma_{\bfR}$		& 	\cellcolor{green!25}$0.041$ 	& 	\cellcolor{green!25}$0.044$			& 	\cellcolor{green!45}$0.041$			&	\cellcolor{blue!25}$0.036$			& \cellcolor{blue!50}$0.031$	& \cellcolor{blue!70}$0.03$	\\\hline %
      \end{tabular} 
    \end{center} 
    \caption{The accuracy of QA under random initial misalignments, for the transformation estimation ("\textit{\textbf{TE}}") and point set alignment ($\boldsymbol{K>1}$).} %
    \label{tab:experiment_type_1} 
\end{table}

\section{Experimental Evaluation}\label{sec:EXPERIMENTS} 

The current generation of D-WAVE annealers does not support the precision of weights in $\bfP$ necessary for our method \cite{DWAVE_QPU}\footnote{the current generation natively supports $9$-bit floating-point numbers}. 
It is foreseeable that future generations will enable a higher accuracy for couplings. 
We thus implement and test QA with an AQC sampler on a conventional computer (Intel \textit{i7-6700K} CPU with $32$GB RAM). 
All quantitative tests are performed with $21$ binary variables corresponding to the size of the $\bfQ$ basis in 2D. 

We report two error metrics, \textit{i.e.,} the alignment error $e_{2D}$ and the transformation discrepancy $e_{\bfR}$, together with their standard deviations denoted by $\sigma_{2D}$ and $\sigma_{\bfR}$, respectively. 
The alignment error $e_{2D} = \frac{\norm{\bfR \bfY - \bfX}_\mathcal{HS}}{\norm{\bfX}_\mathcal{HS}}$ ($\norm{\cdot}_\mathcal{HS}$ denotes the Hilbert-Schmidt norm) measures how accurately the aligned shape coincides with the reference and requires ground truth correspondences. 
The transformation discrepancy is defined as $e_{\bfR} = \norm{\bfI - \bfR \bfR^{\mathsf{T}}}_\mathcal{HS}$, 
where $\bfR$ is the recovered rotation. 
It measures how closely the recovered transformation resembles a valid rigid transformation. 
The usage of two complementary metrics is necessary because a low $e_{\bfR}$ does not automatically imply an accurate registration. 
On the other hand, a low $e_{2D}$ does not quantify how rigid the recovered transformation is. 

\noindent\textbf{Datasets and Proof of Concept.} 
We use four 2D datasets, \textit{i.e.,} \textit{fish} \cite{Myronenko2010}, \textit{qubit}, \textit{kanji} and \textit{composer} with cardinalities varying from $91$ (\textit{fish}) to $7676$ (\textit{composer}), see Fig.~\ref{fig:DIFFERENT_2D_} for qualitative registration results. 
For point sets with up to a few thousand points, the simulation time $\tau_\bfP < 1$ sec. 
For ${\sim}7.7k$, $\tau_\bfP$ grows to $20.178$ sec (by a factor of ${\sim}10^4$). 
Simulation with $n = 30$ takes already ${\sim}2.5$ days. 
More binary variables allow for more elements in the basis $\bfQ$ resulting in more accurate alignment. 
Note that even with $80$ qubits, \textit{i.e.,} for problems with $n = 80$, annealing on AQC takes around $100$ $ms$. 
A simulation with $n = 80$ is not possible even on a conventional supercomputer in a reasonable time. 

\begin{figure}[t!] 
\centering 
\includegraphics[width=1.0\linewidth]{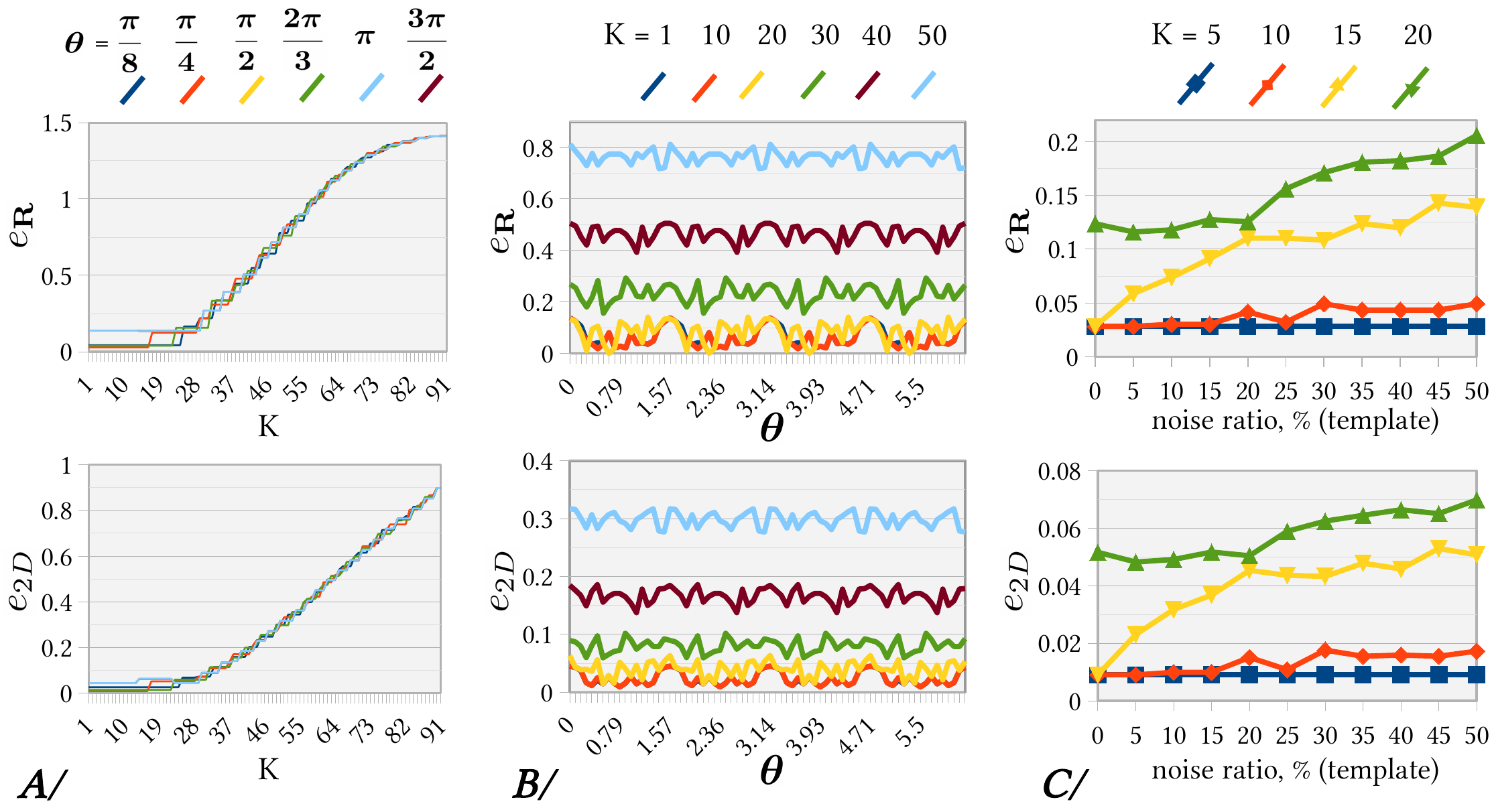} 
\vspace{-15pt} 
\caption{ The metrics as the functions of \textbf{\textit{A/:}} the size of the point interaction region parametrised by $K$;  
\textbf{\textit{B/:}} the angle of initial misalignment $\theta$; \textbf{\textit{C/:}} the template noise ratio. 
} 
\label{fig:drawings_plots} 
\end{figure}

\noindent\textbf{Initial Misalignment and Point Linking.} 
We test how accurately our method recovers the transformation under the random angle of initial misalignment $\theta$ and the different size 
of the point linking region. 
We generate $500$ random transformations in the range $\theta \in [0; 2\pi]$ of the \textit{fish} dataset and resolve them with QA, for each $K \in \{1, 10, 20, 30\}$. 
The results are summarised in Table~\ref{tab:experiment_type_1}. 
We see that $e_{2D}$ correlates with $e_{\bfR}$ for all tested $K$. 
For $K = 30$ --- which corresponds to one third of the template points --- both metrics are still comparably low. 
We also study how the choice of the point interaction region or $K$ affects the accuracy of the transformation recovery and 
plot $e_{2D}$ and $e_{\bfR}$ as the functions of $K$ for several angles of initial misalignment $\theta$ in Fig.~\ref{fig:drawings_plots}-A. 
Interacting points are determined with the $K$ nearest neighbour rule for each $\bfx_n$. 
Recall that according to the singularity theorem \cite{proof_singularity_2019}, the globally multiply-linked alignment 
(here, $K = 91$) results in a shrinkage of the template to a single point, which is observed experimentally. 
Next, we systematically vary the angle of initial misalignment $\theta$ in the range $[0; 2\pi]$ with the angular step $\frac{\pi}{36}$ and report $e_{2D}$ and $e_{\bfR}$ as the functions of $\theta$, for $K \in \{1, 10, 20, 30, 40, 50\}$. 
This test reveals the differences in the transformations caused by $\theta$, which arise due to the composition and the expressiveness of the chosen basis $M$, see Fig.~\ref{fig:drawings_plots}-B. 
QA is almost agnostic to $\theta$, which is a desirable property of every point set alignment method.  

\noindent\textbf{Sensitivity to Noise.} 
We systematically add uniformly distributed noise to the template and test the robustness of the proposed QA to outliers in the data, since real data often contains outliers. 
The highest template noise ratio amounts to $50\%$. 
Each metric for every noise ratio and every $K$ is averaged over $50$ runs, see Fig.~\ref{fig:drawings_plots}-C. 
$\sigma_{\bfR}$ and $\sigma_{2D}$ do not exceed $0.057$ and $0.03$, respectively. 
We observe both the increasing alignment error and the discrepancy in the obtained transformations with the increasing noise level. 
For small $K$, nonetheless, even large noise ratios seem not to influence the metrics significantly. 

\begin{figure}[t!] 
\centering 
\includegraphics[width=0.97\linewidth]{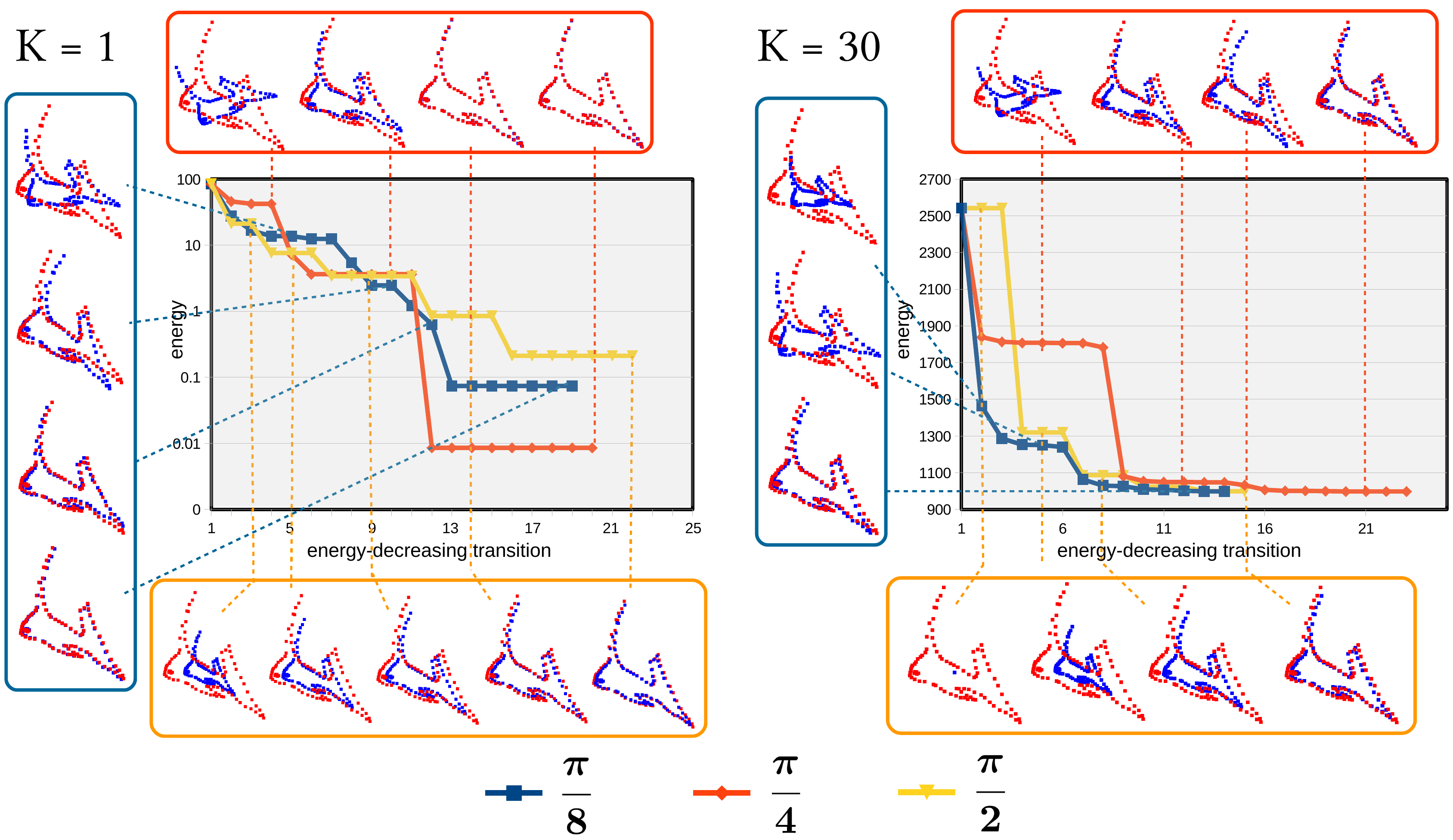} 
\vspace{-5pt} 
\caption{ 
The sequences of energy-decreasing transitions and the corresponding energy values observed in our sampler, for transformation estimation ($K = 1$) and point set alignment with $K = 30$ interactions per $\bfx_n$. 
Besides the graphs, we visualise alignment results for selected energy values and the angle of initial misalignment $\theta \in \big\{\frac{\pi}{8}, \frac{\pi}{4}, \frac{\pi}{2}\big\}$. 
} 
\label{fig:energy_decreasing_transitions} 
\end{figure}

\noindent\textbf{Spectral Gap Analysis.} 
Spectral gap $\Delta(\Ham)$ is the difference between the energy of the ground state and the second-lowest eigenstate. 
Each problem has an intrinsic and unique $\Delta(\Ham)$. 
Even though a rigorous analysis of the spectral gap is out of the scope of this paper, we make several qualitative observations about the energy landscape of QA, the difference in the energy values and the corresponding registrations for one exemplary problem. 
In Fig.~\ref{fig:energy_decreasing_transitions}, we plot the sequences of energy-decreasing transitions together with the energy values in the experiment with
\textit{fish}, for three $\theta$ values. %
We notice that some solutions have very small differences in the energies and are qualitatively indistinguishable from each other. 
This is accounted for by the choice of the additive basis, \textit{i.e.,} that the same alignment can be encoded in different ways. 
In contrast, we see significant differences in the energy values of the qualitatively different solutions (orders of magnitudes larger in the analysed experiment). 
We conclude that even though $\Delta(\Ham)$ is small, the alignments corresponding to several few lowest eigenstates are qualitatively similar. 
This suggests that our selection of the basis leads to problems with sufficient spectral gaps. 

\section{Conclusions}\label{sec:conclusion} 

This paper introduces AQC for the computer vision community and shows that fundamental low-level problems can be brought to a representation suitable for solving on AQC.  
In simulations on a classical computer and in a wide range of scenarios, our QA is shown to successfully recover 2D transformations which are close approximations of globally optimal transformations. %
With the chosen basis of $20$ elements, the solutions result in low transformation discrepancy and alignment errors. 
Observations on how to avoid singularities as well as the noise sensitivity and  spectal gap analysis complement the experimental section. 

In future work, our technique can be extended to affine transformations and other related computer vision problems. 
We hope to see more research on computer vision methods with quantum hardware in the next decades. 
\noindent\textbf{\small Acknowledgements.} 
{\small 
This work was supported by the ERC Consolidator Grant 770784. 
VG is grateful to Polina Matveeva for many enlightening discussions on the physical foundations of adiabatic quantum computing. 
The authors thank Bertram Taetz and Hanno Ackermann for reviewing an earlier version of this paper. 
}

\appendix

\section{Appendix}\label{app:appendix} 

\renewcommand{\thefigure}{\Roman{figure}}
\setcounter{figure}{0}

\renewcommand{\theequation}{\roman{equation}}
\setcounter{equation}{0}

In this additional section, we provide details on the selection of the annealing rate, analyse the structure of $\bfP$ and formalise the unembedding, \textit{i.e.,} the conversion of the solution to QUBOP \eqref{eq:xtPx} to the solution of the original alignment problem on point sets. 
We preserve the notations referring to the sections and equations from the main matter. 
The equations and the figure introduced in this supplement are equipped with Roman numerals. 
\noindent\textbf{Annealing Rate.} Suppose $E_n(s)$ is the ground state of instantaneous Hamiltonian, 
$E_n(0)$ is the initial state (ground state) of the system and 
$E_m(s)$ is any other excited state of the instantaneous Hamiltonian. 
Let $s = \frac{t}{T} \in [0; 1]$, where $T$ is the overall time of interpolation and $t$ is physical time. 
Then, according to \cite{Amin2009}, $T$ has to be chosen so that  
\begin{equation} 
 T \gg \frac{|\langle E_m(s) | d H / d s | E_n(s) \rangle |}{E_{nm}(s)^2 }, \; \forall m \ne n, 
\end{equation} 
where $d H / d s$ is the rate of change of Hamiltonian with respect to $s$ and $E_{nm}$ is the difference in the corresponding instantaneous energies. 

\begin{figure}[t!] 
\centering 
\includegraphics[width=1.0\linewidth]{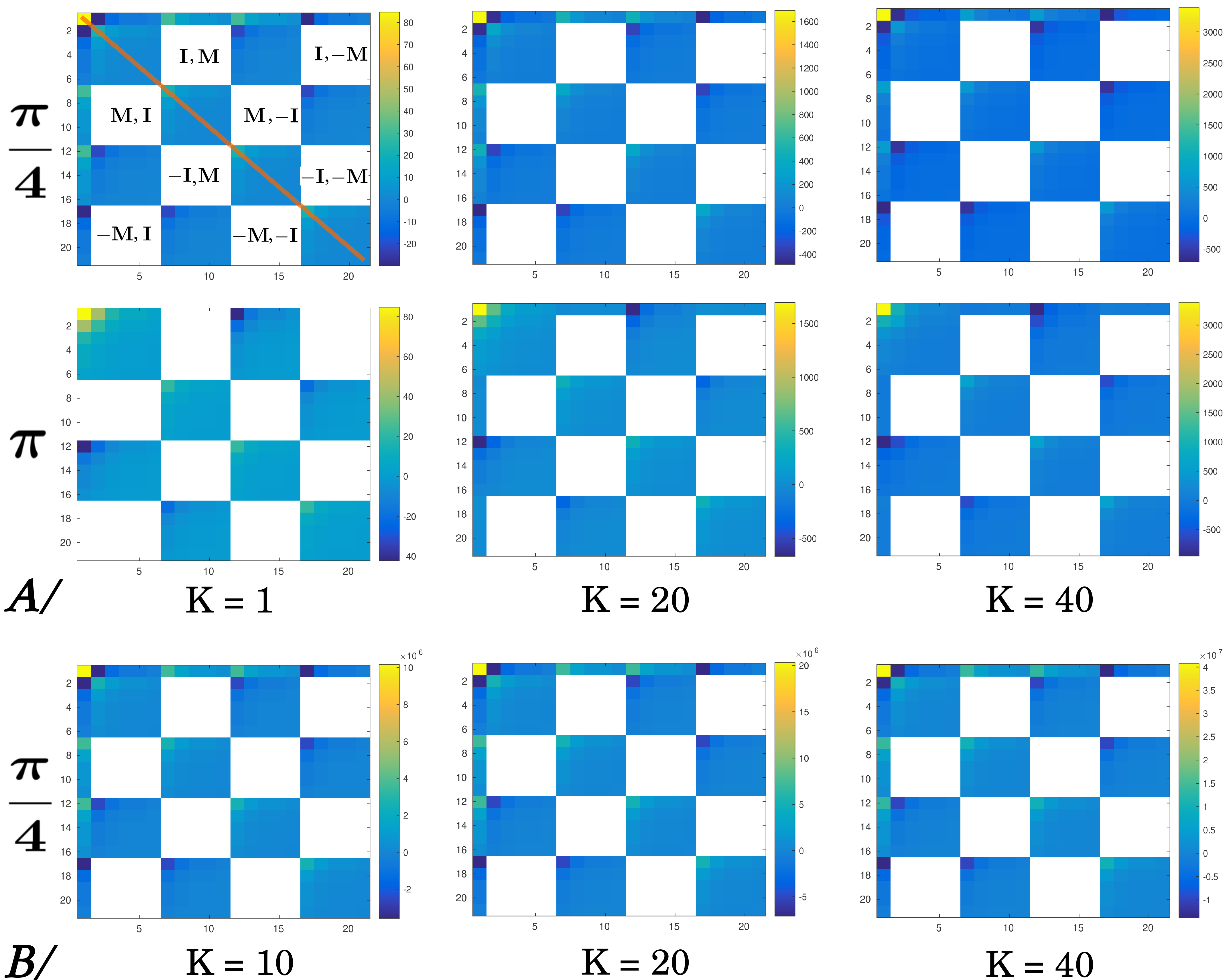} 
\caption{Exemplary visualisations of the weight matrix $\bfP = \boldsymbol{\Phi} \boldsymbol{\Phi}^\mathsf{T}$ in the experiment with clean (\textbf{\textit{A/}}) and noisy data with $35\%$ of outliers in the template (\textbf{\textit{B/}}), for $K \in \{1, 10, 20, 40\}$ and $\theta \in \big\{\frac{\pi}{4}, \pi\big\}$. 
The colour scheme and the range of energy values are given to the right of each $\bfP$. 
White colour stands for zero entries. 
The diagonal values in $\bfP$ represent biases (marked in orange on the top left), and non-zero elements represent couplings between the qubits. 
In the visualisation on the top left, we list the pairs of $\bfC \in \{\bfI, \bfM, -\bfI, -\bfM\}$ eventually leading to zero matrices. 
} 
\label{fig:app1} 
\end{figure}

\noindent\textbf{Analysis of $\bfP$}. 
Fig.~\ref{fig:app1} visualises several exemplary weight matrices $\bfP$ from the experiments with clean and noisy data (see Sec.~\ref{sec:EXPERIMENTS}). 
There are several observations. 
First, $\bfP = \boldsymbol{\Phi} \boldsymbol{\Phi}^\mathsf{T}$ is symmetric upon algorithm design. 
We also see that the columns of $\boldsymbol{\Phi}$ can be arbitrarily reshuffled as long as the correspondences are preserved\footnote{a reshuffling of rows requires changing the order of elements in $\bfQ$}. 
Second, $\bfP$ contains regularly arranged zero submatrices, due to our choice of the basis. 
As soon as a row of $\boldsymbol{\Phi}$ induced by $q \bfC_{\bfI}$, where $\bfC_{\bfI} \in \{\bfI, -\bfI\}$, is multiplied by a column of $\boldsymbol{\Phi}^\mathsf{T}$ induced by $q \bfC_{\bfM}$, where $\bfC_{\bfM} \in \{\bfM, -\bfM\}$, and vice versa, we obtain a zero entry in $\bfP$. 
The reason is that 
\begin{equation} 
\begin{cases}
 [ \bfI \, \sum_i \bfy_i ]^\mathsf{T}  &[\bfM \, \sum_j \bfy_j]\;\;\, = 0\\ 
 [ -\bfI \, \sum_i \bfy_i ]^\mathsf{T} &[\bfM \, \sum_j \bfy_j]\;\;\, = 0\\ 
 [ \bfI \, \sum_i \bfy_i ]^\mathsf{T}  &[-\bfM \, \sum_j \bfy_j] = 0\\ 
 [ -\bfI \, \sum_i \bfy_i ]^\mathsf{T} &[-\bfM \, \sum_j \bfy_j] = 0
\end{cases}, 
\end{equation} 
if $\sum_i \bfy_i = \sum_j \bfy_j$, which holds in our case since each row of $\boldsymbol{\Phi}$ except the first row includes all points of $\bfY$ multiplied by a single basis element $\bfQ_k$ (see Fig.~\ref{fig:app1}-(top left) for $\bfC$ pairs resulting in zero matrices). 
Third, the structure of $\bfP$ reflects that its diagonal entries encode biases, and non-diagonal elements represent couplings between the qubits. 
With the increasing $K$, the span of the absolute energy values increases, due to the higher number of point interactions. 
As expected, $\bfP$ depends on data and the angle of initial misalignment between the point sets. 
For all possible inputs and initial conditions --- point sets of different cardinalities, $K$ and $\theta$ --- the structure of $\bfP$ is the same for the chosen basis. 
From $\bfP$, we also recognise that the considered alignment problem is not purely combinatorial and requires high-precision weights $J_{j,k}$ in \eqref{eq:Ising_Model_Hamiltonian}. 

\noindent\textbf{Unembedding.} 
Unembedding is the decoding of the solution to QUBOP \eqref{eq:xtPx} to the solution of the original alignment problem. 
Upon the design, our QA method assembles the entries of the transformation matrix in the additive basis $\bfQ_k$ (see  Secs.~\eqref{sec:2D_transformation_estimation}--\eqref{ssec:particle_dynamics_based_alignment}). 
Suppose $\hat{\bfq}$ is the measurement result of $\bfq$, \textit{i.e.,} it is a classical bitstring with $K+1$ elements. 
Recall that $\bfq_1$ is reserved for reference points and does not contribute to the assembly of the transformation. 
Once $\hat{\bfq}$ is measured and returned, we obtain the corresponding transformation $\bfR$ by summing up $\bfQ_k$ multiplied by $\hat{\bfq}_{k+1}$: 
\begin{equation} 
  \bfR = \sum_k \hat{\bfq}_{k+1} \bfQ_k. 
\end{equation} 
The obtained $\bfR$ is an affine transformation. 
If the solution has to represent a valid rotation matrix $\bfR_{\text{r}}$, $\bfR$ can be projected to the rotation group by solving the 
\textit{closest orthogonal approximation problem with constraints}: 
\begin{equation}\label{eq:closest_rotation_matrix} 
\begin{aligned} 
  & \;\;\;\;\;\;\, \min \norm{\bfR_{\text{r}} - \bfR}_\mathcal{HS}^2, \\ 
  & \text{s.~t.~} \; \bfR_{\text{r}}^{-1} = \bfR_{\text{r}}^\mathsf{T} \;\text{and}\; \operatorname{det}(\bfR_{\text{r}}) = 1. 
\end{aligned} 
\end{equation} 
For a solution to \eqref{eq:closest_rotation_matrix} by singular value decomposition, see \cite{Higham1989}. %

{\small 
\bibliographystyle{ieee_fullname} 
\bibliography{egbib} 
} 

\end{document}